\documentclass{itatnew}
\usepackage{todonotes}
\usepackage{soul}
\usepackage[numbers]{natbib}
\presetkeys{todonotes}{inline}{}
\presetkeys{todonotes}{prepend}{}
\presetkeys{todonotes}{caption=TODO}{}

\definecolor{darkgreen}{rgb}{0.0, 0.5, 0.0}

\usepackage[normalem]{ulem}
\def\ODdel#1{\bgroup\markoverwith{\textcolor{darkgreen}{\rule[0.5ex]{2pt}{1pt}}}\ULon{#1}}

\def\area#1{{\color{darkgreen}area:\it #1}}
\def\food#1#2{{Dial. state #1: \color{blue}food:\it #2}}
\def\pricerange#1{{\color{orange}pricerange:\it #1}}
\def\sys#1{{\color{purple}System: \it #1}}
\def\usr#1{{\color{brown}User: \it #1}}

\begin{document}

\title{Recurrent Neural Networks for Dialogue State Tracking}

\author{Ondřej Plátek \and Petr Bělohlávek \and Vojtěch Hudeček  \and Filip Jurčíček}

\institute{Charles University in Prague, Faculty of Mathematics and Physics \\
\email{\{oplatek,jurcicek\}@ufal.mff.cuni.cz},\\
\email{me@petrbel.cz},\\
\email{vojta.hudecek@gmail.com},\\
\texttt{http://ufal.mff.cuni.cz/ondrej-platek}}

\maketitle              



\begin{abstract}
This paper discusses models for dialogue state tracking using recurrent neural networks (RNN).
We present experiments on the standard dialogue state tracking (DST) dataset, DSTC2~\cite{henderson2014second}.
On the one hand, RNN models became the state of the art models in DST,
on the other hand, most state-of-the-art DST models are only turn-based and require dataset-specific preprocessing (e.g. DSTC2-specific) in order to achieve such results.
We implemented two architectures which can be used in an incremental setting and require almost no preprocessing.
We compare their performance to the benchmarks on DSTC2 and discuss their properties.
With only trivial preprocessing, the performance of our models is close to the state-of-the-art results.\footnote{
    {\bf Acknowledgment:} We thank Mirek Vodolán and Ondřej Dušek for useful comments.
    This research was partly funded by the Ministry of Education, Youth and Sports of the Czech Republic under the grant agreement LK11221, core research funding, grant GAUK 1915/2015, and also partially supported by SVV project number 260 333. 
    We gratefully acknowledge the support of NVIDIA Corporation with the donation of the Tesla K40c GPU used for this research.
    Computational resources were provided by the CESNET LM2015042 and the CERIT Scientific Cloud LM2015085, provided under the programme ``Projects of Large Research, Development, and Innovations Infrastructures''.
    }
\end{abstract}

\section{Introduction}
 
Dialogue state tracking (DST) is a standard and important task for evaluating task-oriented conversational agents~\cite{williams2013dialog, henderson2014second, henderson2014third}.
Such agents play the role of a domain expert in a narrow domain, and users ask for information through conversation in natural language (see the example system and user responses in~Figure~\ref{fig:example}).
A dialogue state tracker summarizes the dialogue history and maintains a probability distribution over the (possible) user's goals (see annotation in~Figure~\ref{fig:example}).
Dialogue agents as introduced in~\cite{young2010hidden} decide about the next action based on the dialogue state distribution given by the tracker.
User's goals are expressed in a~formal language, typically represented as a dialogue act items (DAIs) (see Section~\ref{sec:dst}) and the tracker updates probability for each item.
The dialogue state is a latent variable~\cite{young2010hidden} and one needs to label the~conversations in order to train a dialogue state tracker using supervised learning.
It was shown that with a better dialogue state tracker, conversation agents achieve better success rate in overall completion of the their task~\cite{jurvcivcek2012reinforcement}.

\begin{figure}
   \dots \\
    \food{n}{None}, \area{None}, \pricerange{None} \\
    \sys{What part of town do you have in mind?} \\
    \usr{West part of town.} \\
    \food{n+1}{None}, \area{west}, \pricerange{None} \\
    \sys{What kind of food would you like?} \\
    \usr{Indian} \\
    \food{n+2}{Indian}, \area{west}, \pricerange{None} \\
    \sys{India House is a nice place in the west of town serving tasty Indian food.} \\
    \dots
    \caption{Example of golden annotation of Dialogue Act Items (DAIs). The dialogue act items comprise from act type (all examples have type {\it inform}) and slots ({\it food, area, pricerange}) and their values (e.g. {\it Indian, west, None}).}
\vspace{-0.70em}
\label{fig:example}
\end{figure}

This paper compares two different RNN architectures for dialogue state tracking (see Section~\ref{sec:model}).
We describe state-of-the art word-by-word dialogue state tracker architectures and propose to use a new encoder-decoder architecture for the DST task (see Section~\ref{sec:eval}).

We focus only on the {\it goal} slot predictions because the other groups are trivial to predict\footnote{The slots {\it Requested} and {\it Method} have accuracies 0.95 and 0.95 on the test set according to the state-of-the-art~\cite{williams2014web}.}. 

We also experiment with re-splitting of the DSTC2 data because there are considerable differences between the standard train and test datasets~\cite{henderson2014second}.
Since the training, development and test set data are distributed differently, the resulting performance difference between training and test data is rather high.
Based on our experiments, we conclude that DSTC2 might suggest a~too pessimistic view of the state-of-the-art methods in dialogue state tracking caused by the data distribution mismatch.

\section{Dialogue state tracking on DSTC2 dataset}\label{sec:dst}
Dialogue state trackers maintain their believes beliefs about users' goals by updating probabilities of dialogue history representations.
In the DSTC2 dataset, the history is captured by dialogue act items and their probabilities.
A Dialogue act item is a triple of the following form $(actionType, slotName, slotValue)$.

The DSTC2 is a standard dataset for DST, and most of the state-of-the-art systems in DST have reported their performance on~this dataset~\cite{henderson2014second}. 
The full dataset is freely available since January 2014 and it contains 1612 dialogues in the training set, 506 dialogues in the development set and 1117 dialogues in the test set.\footnote{Available online at~\url{http://camdial.org/~mh521/dstc/}.}
The~conversations are manually annotated at the turn level where the hidden information state is expressed in form of $(actionType, slotName, slotValue)$ based on the~domain ontology.
The task of the domain is defined by a~database of restaurants and their properties\footnote{There are six columns in the database: name, food, price\_range, area, telephone, address.}.
The~database and the~manually designed ontology that captures a~restaurant domain are both distributed with the~dataset.

\section{Models}\label{sec:model}
Our models are all based on a RNN encoder~\cite{werbos1990backpropagation}. 
The models update their hidden states $h$ after processing each word similarly to the RNN encoder of~\citet{zilka2015incremental}.
The encoder takes as inputs the previous state $h_{t-1}$, representing history for first $t-1$ words, and features $X_t$ for the current word $w_t$. 
It outputs the current state $h_t$ representing the whole dialogue history up to current word.
We use a~Gated Recurrent Unit (GRU) cell~\cite{cho2014gru} as the update function instead of a simple RNN cell because it does not suffer from the vanishing gradient problem~\cite{hochreiter2001gradient}.
The model optimizes its parameters including word embeddings~\cite{bengio2003neural} during training.

For each input token, our RNN encoder reads the word embedding of this token along with several binary features. 
The binary features for each word are:
\begin{itemize}
	\item the speaker role, representing either user or system,
    \item and also indicators describing whether the word is part of a named entity representing a value from the database.
\end{itemize}
Since the DSTC2 database is a simple table with six columns, we introduce six binary features firing if the word is a substring of named entity from the given column.
For example, the word {\it indian} will not only trigger the feature for column $food$ and its value {\it indian} but also for column restaurant $name$ and its value {\it indian heaven}.
The features make the data dense by abstracting the meaning from the lexical level. 

Our model variants differ only in the way they predict {\it goal} labels, i.e., $food$, $area$ and $price range$ from the RNN's last encoded state.\footnote{Accuracy measure with schedule 2 on slot $food$, $area$ and $price range$ about which users can inform the system is a featured metric for DSTC2 challenge~\cite{henderson2014second}.} 
The first model predicts the output slot labels independently by employing three independent classifiers (see Section~\ref{sec:indep}).
The second model uses a decoder in order to predict values one after the other from the $h_{T}$ (see Section~\ref{sec:encdec}).

The models were implemented using the~TensorFlow~\cite{abaditensorflow} framework. 

\begin{figure}
\vspace{-0.80em}
\begin{center}
\includegraphics[height=9em]{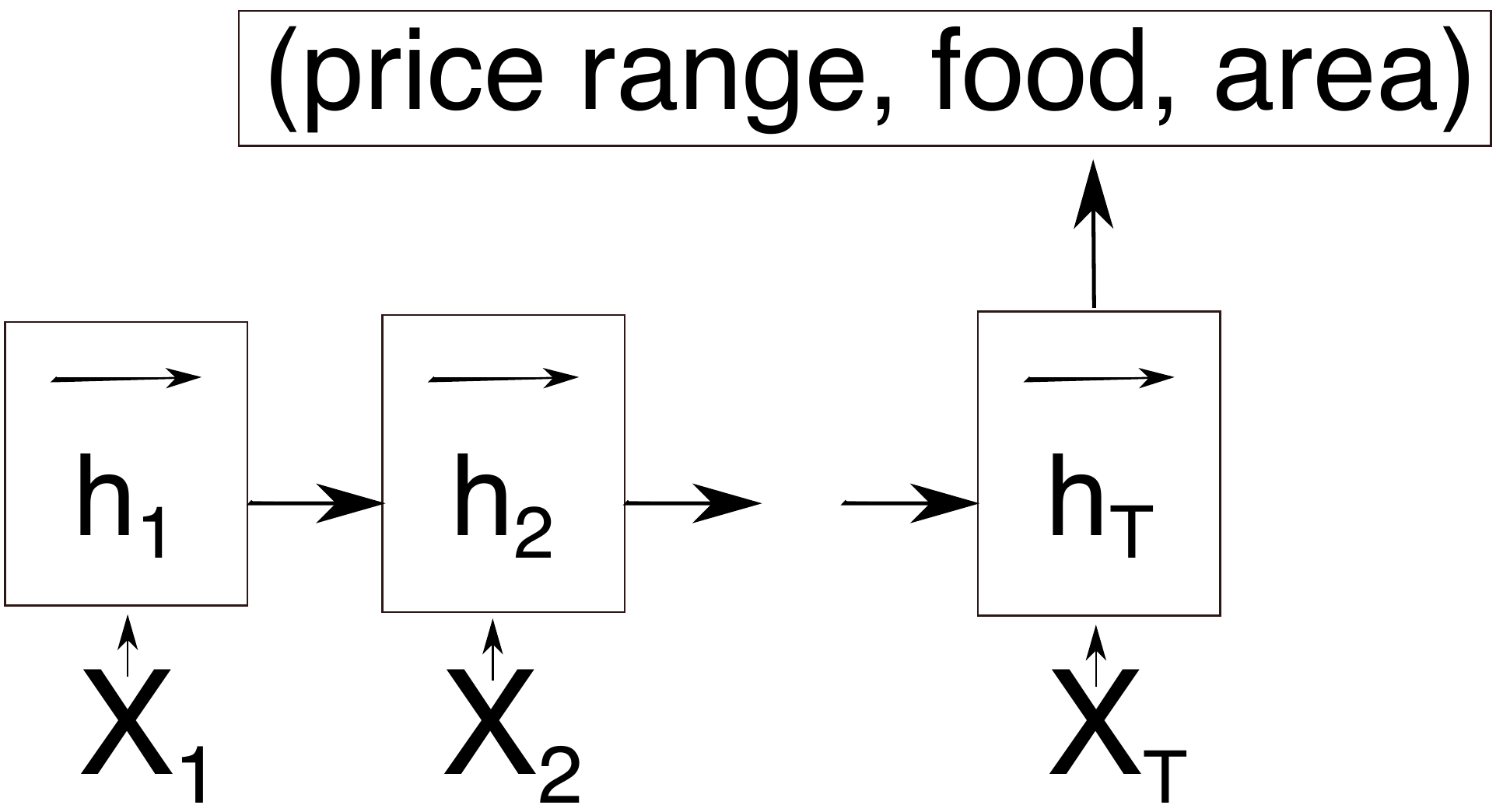}
\caption{The joint label predictions using RNN from last hidden state $h_T$. The $h_T$ represents the whole dialog history of $T$ words. The RNN takes as input for each word $i$ an embedding and binary features concatenated to~vector~$X_{i}$.}
\end{center}
\vspace{-0.70em}
\label{fig:encjoint}
\end{figure}

\subsection{Independent classifiers for each label}
\label{sec:indep}
\begin{figure}
\begin{center}
\includegraphics[height=9.5em]{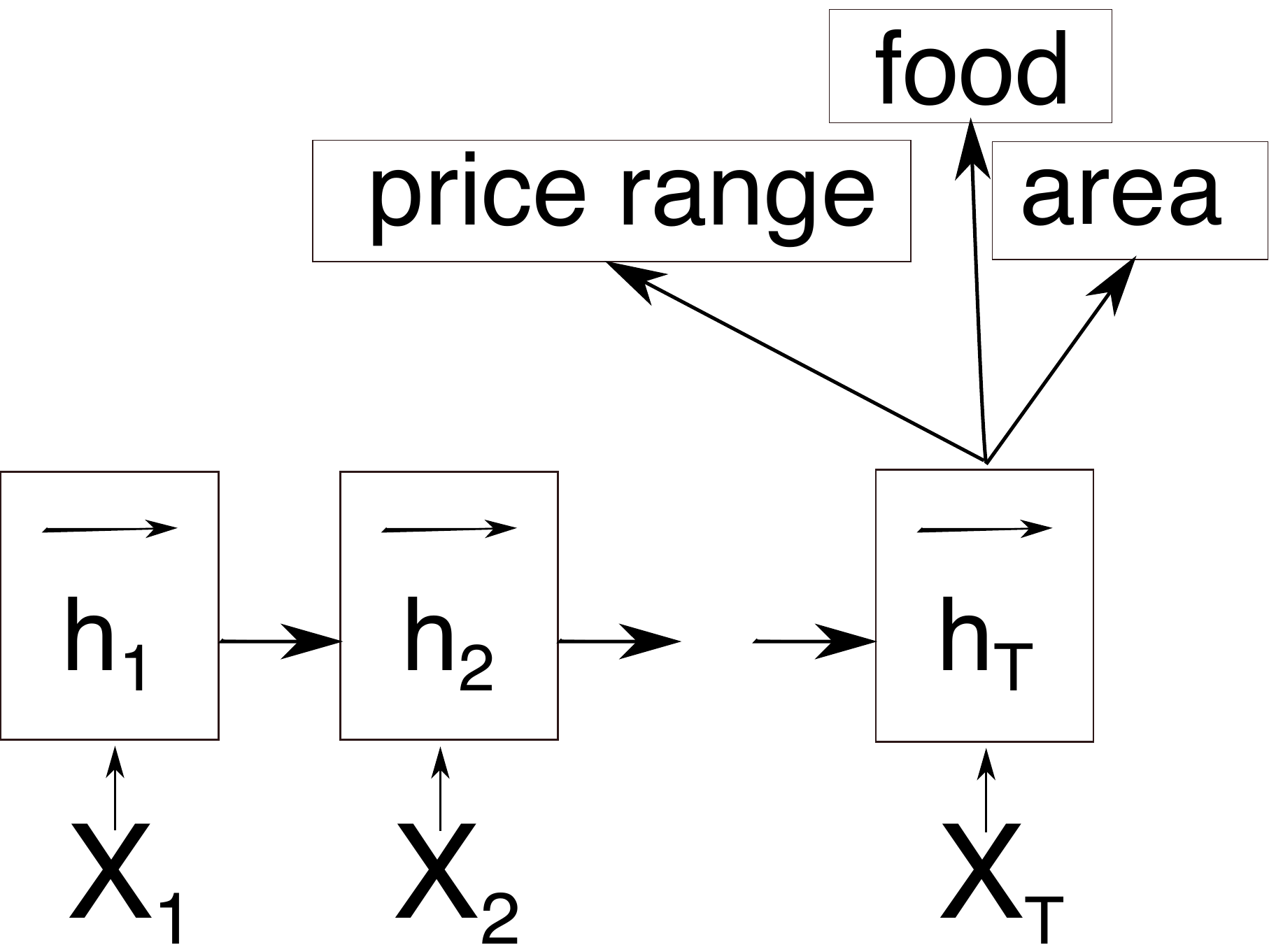}
\caption{The RNN encodes the word history into dialogue state $h_T$ and predicts slot values independently.}
\end{center}
\vspace{-0.80em}
\label{fig:encind}
\end{figure}

The independent model (see Figure~\ref{fig:encind}) consists of three models which predict $food$, $area$ and $price range$ based on the last hidden state $h_{T}$ independently.
The~independent slot prediction that uses one classifier per slot is straightforward to implement, but the model introduces an unrealistic assumption of uncorrelated slot properties.
In case of DSTC2 and the Cambridge restaurant domain, it is hard to believe that, e.g., the slots $area$ and $price range$ are not correlated.

We also experimented with a~single classifier which predicts the labels jointly (see Figure~\ref{fig:encjoint}) but it suffers from data sparsity of the predicted tuples, so we focused only on the independent label prediction and encoder-decoder models.

\subsection{Encoder-decoder framework}
\label{sec:encdec}
\begin{figure}
\includegraphics[width=0.5\textwidth]{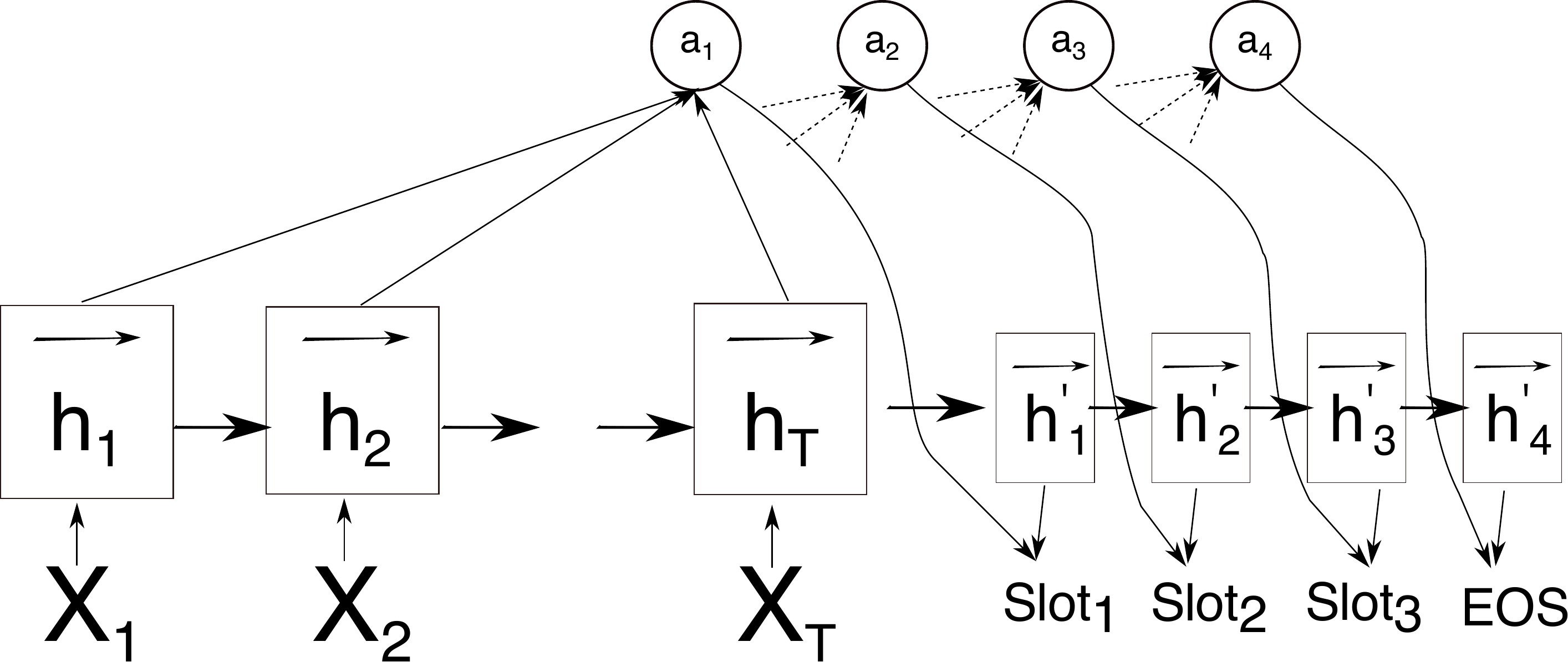}
\caption{Encoder decoder with attention predicts goals.}
\label{fig:encdec}
\end{figure}

We cast the slot predictions problem as a sequence-to-sequence predictions task and we use a~encoder-decoder model with attention~\cite{bahdanau2014neural} to learn this representation together with slot predictions (see Figure~\ref{fig:encdec}).
To our knowledge, we are the~first who used this model for dialogue state tracking.
The model is successfully used in machine translation where it is able to handle long sequences with good accuracy~\cite{bahdanau2014neural}.
In DST, it captures correlation between the decoded slots easily. 
By introducing the encoder-decoder architecture, we aim to overcome the data sparsity problem and the~incorrect independence assumptions.

We employ the encoder RNN cell that captures the history of the~dialogue which is represented as a~sequence of words from the~user and the~system.
The words are fed to the encoder as they appear in the dialogue - turn by turn - where the user and the system responses switch regularly.
The encoder updates its internal state $h_T$ after each processed word.
The RNN decoder model is used when the system needs to generate its output, in our case it is at the end of the user response.
The decoder generates arbitrary length sequences of words given the encoded state $h_T$ step by step.
In each step, an output word and a~new hidden state $h_{T+1}$ is generated.
The generation process is finished when a special token End of Sequence (EOS) is decoded.
This mechanism allows the model to terminate the output sequence.
The attention part of model is used at decoding time for weighting the importance of the history. 

The disadvantage of this model is its complexity.
Firstly, the model is not trivial to implement\footnote{We modified code from the TensorFlow `seq2seq` module.}. 
Secondly, the decoding time is asymptotically quadratic in the length of the decoded sequences, but our target sequences are always four tokens long nevertheless.

\section{Experiments}\label{sec:exp}
The results are reported on the standard DSTC2 data split where we used 516 dialogues as a validation set for early stopping~\cite{prechelt1998early} and the remaining 1612 dialogues for training.
We use 1-best Automatic Speech Recognition (ASR) transcriptions of conversation history of the input and measure the~joint slot accuracy.
The models are evaluated using the recommended measure accuracy~\cite{henderson2014second} with schedule 2 which skips the first turns where the believe tracker does not track any values. 
In addition, our models are also evaluated on a randomly split DSTC2 dataset (see Section~\ref{sec:split}).

For all our experiments, we train word embeddings of size 100 and use the encoder state size of size 100, together with a dropout keep probability of $0.7$ for both encoder inputs and outputs.
These parameters are selected by a grid search over the hyper-parameters on development data.

\subsection{Training}
\label{sec:train}
The training procedure minimizes the cross-entropy loss function using the Adam optimizer~\cite{kingma2014adam} with a batch size of 10.
We train predicting the~goal slot values for each turn.
We treat each dialogue turn as a separate training example, feeding the whole dialogue history up to the current turn into the encoder and predicting the slot labels for the current turn.

We use early stopping with patience~\cite{prechelt1998early}, validating on the development set after each epoch and stopping if the three top models does not change for four epochs.

The predicted labels in DST task depend not only on the last turn, but on the dialogue full history as well.
Since the lengths of dialogue histories vary a lot\footnote{The maximum dialogue history length is 487 words and 95\% percentile is 205 words for the training set.} and we batch our inputs, we separated the dialogues into ten buckets accordingly to their lengths in order to provide a computational speed-up. We reshuffle the data after each epoch only within each bucket.

In informal experiments, we tried to speed-up the training by  optimizing the parameters only on the last turn\footnote{The prediction was conditioned on the full history but we back-propagated the error only in words within the last turn.} but the performance dropped relatively by more than 40\%.

\subsection{Comparing models}
\label{sec:eval}

\begin{figure}
\vspace{-0.80em}
    \begin{center}
\includegraphics[width=0.4\textwidth]{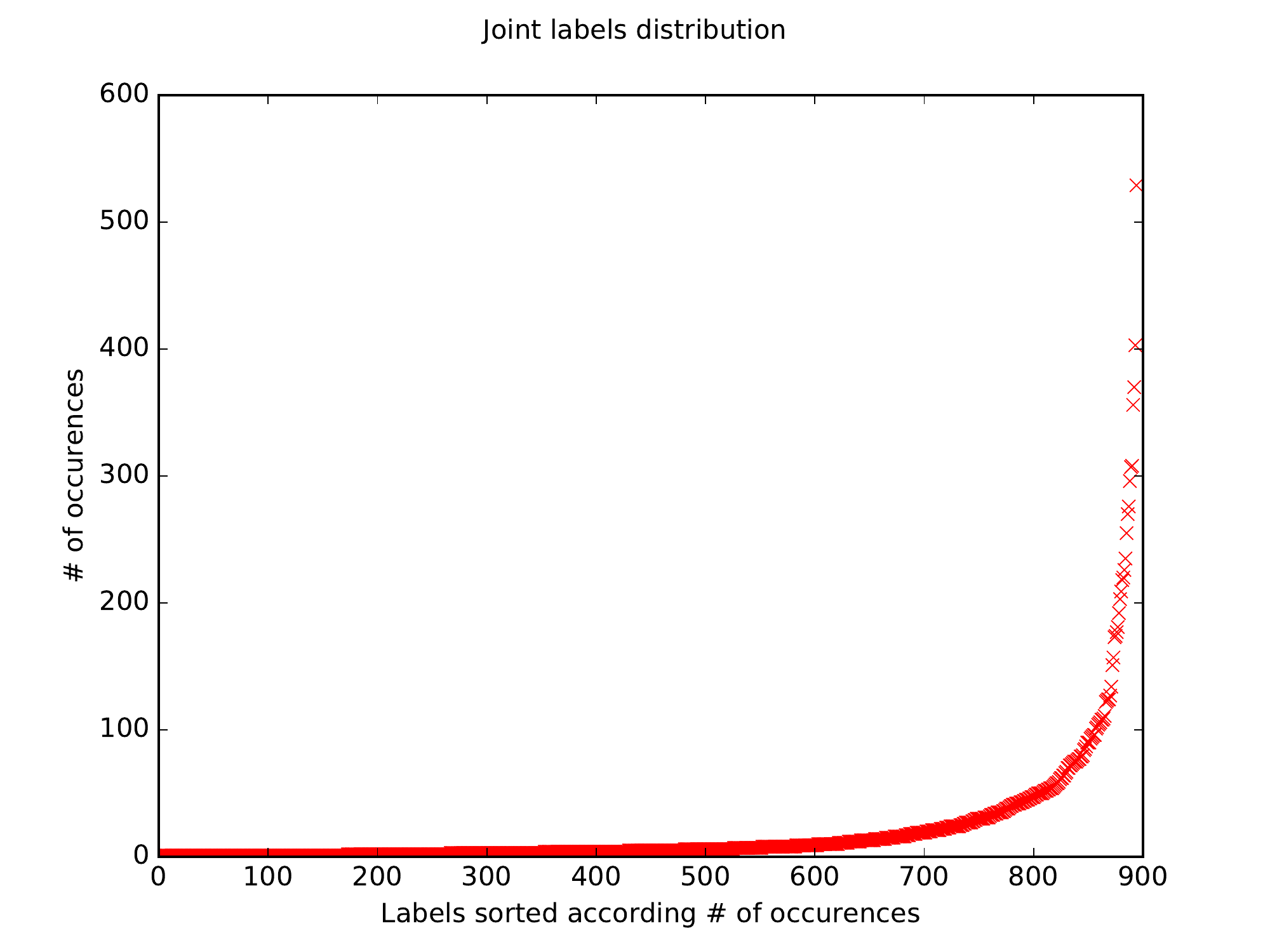}
    \end{center}
\vspace{-1.80em}
\caption{The number occurrences of labels in form of $(food, area, pricerange)$ triples from the least to the most frequent.}
\label{fig:labels}
\end{figure}

Predicting the labels jointly is quite challenging because the distribution of the labels is skewed as demonstrated in~Figure~\ref{fig:labels}.
Some of the labels combinations are very rare, and they occur only in the development and test set so the joint model is not able to predict them.
During first informal experiments the~joint model performed poorly arguably due to data sparsity of slot triples. We further focus on model with independent classifiers and encoder-decoder architecture.

The model with independent label prediction is a strong baseline which was used, among others, in work of~\citet{zilka2015incremental}.
The model suffers less from dataset mismatch because it does not model the correlation between predicted labels.
This property can explain a smaller performance drop between the test set from reshuffled data and the official test set in comparison to encoder-decoder model.

\begin{table}
\vspace{-0.80em}
\begin{center}
\begin{tabular}{l@{\quad}rll}
\hline
\multicolumn{1}{l}{\rule{0pt}{12pt}
                   Model}&\multicolumn{1}{l}{Dev set}&\multicolumn{2}{l}{Test set}\\[2pt]
\hline\rule{0pt}{12pt}
    Indep  &   0.892 & 0.727 \\
    EncDec &   0.867 & 0.730 \\
\hline
    \citet{vodolan2015hybrid} & - & 0.745 \\
    \citet{zilka2015incremental} & 0.69 & 0.72 \\
    \citet{henderson2013deep} & - & 0.737 \\
\hline
    DSTC2 stacking ensemble~\cite{henderson2014second} & - & 0.789 \\
\hline
\end{tabular}
\caption{Accuracy on DSTC2 dataset. The first group contains our systems which use ASR output as input, the second group lists systems using also ASR hypothesis as input. The third group shows the results for ensemble model using ASR output nd also live language understanding annotations.}
\vspace{-2em}
\end{center}
\label{tab:dstc}
\end{table}

Since the encoder-decoder architecture is very general and can predict arbitrary output sequences, it also needs to learn how to predict only three slot labels in the correct order.
It turned out that the architecture learned to predict quadruples with three slot values and the EOS symbol quickly, even before seeing a~half of the training data in the first epoch.\footnote{We could have modified the decoder to always predict three symbols for our three slots, but our experiments showed that the encoder-decoder architecture does not make mistakes at predicting the order of the three slots and EOS symbol.}  
At the end of the first epoch, the system made no more mistakes on predicting slot values in the incorrect order.
The encoder-decoder system is competitive with state-of-the art architectures and the time needed for learning the output structure was surprisingly short.\footnote{The best model weights were found after 18 to 23 epochs for all model architectures.}

\subsection{Data preparation experiments}
\label{sec:split}

\begin{table}
\vspace{-0.20em}
\begin{center}
\begin{tabular}{l@{\quad}rll}
\hline
\multicolumn{1}{l}{\rule{0pt}{12pt}
                   Model}&\multicolumn{1}{l}{Dev set}&\multicolumn{2}{l}{Test set}\\[2pt]
\hline\rule{0pt}{12pt}
Indep  &   0.87 & 0.89 \\
EncDec &   0.94 & 0.91 \\
\hline
\end{tabular}
\caption{Accuracy of our models on the re-split DSTC2 data.}
\vspace{-2em}
\end{center}
\label{tabsplit}
\end{table}

The data for the DSTC2 test set were collected using a different spoken dialogue system configuration than the data for the validation and the training set.\cite{henderson2014second}.
We intended to investigate the influence of the complexity of the task, hence we merged all DSTC2 data together and created splits of 80\%, 10\% and 10\% for the training, development and test sets.
The~results in Table~2 show that the complexity of the task dropped significantly.

\section{Related work}\label{sec:related}
Since there are numerous systems which reported on the DSTC2 dataset, we discuss only the systems which use RNNs.
In general, the RNN systems achieved excellent results.

Our system is related to the RNN tracker of~\citet{zilka2015incremental},
which reported near state-of-the art results on the~DSTC2 dataset and introduced the first incremental system which was able to update the dialogue state word-by-word with such accuracy.
In contrast to work of~\citet{zilka2015incremental}, we use no abstraction of slot values. 
Instead, we add the additional features as described in Section~\ref{sec:model}.
The first system which used a neural network for dialogue state tracking~\cite{henderson2013deep} used a feed-forward network and more than 10 manually engineered features across different levels of abstraction of the user input, including the~outputs of the spoken language understanding component (SLU).
In our work, we focus on simplifying the architecture, hence we used only features which were explicitly given by the dialogue history word representation and the database.

The system of~\citet{henderson2014word} achieves the~state-of-the-art results and, similarly to our system, it predicts the dialogue state from words by employing a RNN.
On the other hand, their system heavily relies on the user input abstraction.
Another dialogue state tracker with LSTM was used in the~reinforcement setting but the authors also used information from the SLU pipeline~\cite{lee2016dialog}.

An interesting approach is presented in the work of~\citet{vodolan2015hybrid}, who combine a rule-based and a machine learning based approach.
The handcrafted features are fed to an LSTM-based RNN which performs a dialog-state update.
However, unlike our work, their system requires SLU output on its input.

It is worth noting that there are first attempts to train an end-to-end dialogue system even without explicitly modeling the dialogue state~\cite{bordes2016learning}, which further simplifies the architecture of a dialogue system.
However, the reported end-to-end model was evaluated only on artificial dataset and cannot be compared to DSTC2 dataset directly.

\section{Conclusion}\label{sec:conc}
We presented and compared two dialogue state tracking models which are based on state-of-the-art architectures using recurrent neural networks.
To our knowledge, we are the first to use an encoder-decoder model for the dialogue state tracking task, and we encourage others to do so because it is competitive with the standard RNN model.\footnote{The presented experiments are published at \url{https://github.com/oplatek/e2end/} under Apache license. Informal experiments were conducted during the Statistical Dialogue Systems course at Charles University (see \url{https://github.com/oplatek/sds-tracker}).}
The models are comparable to the~state-of-the-art models.  

We evaluate the models on DSTC2 dataset containing  task-oriented dialogues in the~restaurant domain. 
The models are trained using only ASR 1-best transcriptions and task-specific lexical features defined by the task database.
We observe that  dialogue state tracking on DSTC2 test set is notoriously hard and that the task becomes substantially easier if the data is reshuffled.

As future work, we plan to investigate the~influence of the introduced database features on models accuracy.
To our knowledge there is no dataset which can be used for evaluating incremental dialogue state trackers, so it would be beneficial to collect the word-level annotations so one can evaluate incremental DST models.

\bibliographystyle{plainnat}
\bibliography{samplearticle}
\end{document}